\documentclass[runningheads]{llncs}
\usepackage[T1]{fontenc}
\usepackage{graphicx}
\usepackage{amsmath}
\usepackage{amssymb}
\usepackage{algorithm}
\usepackage{algorithmic}
\usepackage{booktabs}
\usepackage{multirow}
\usepackage[table]{xcolor}
\usepackage{arydshln}
\usepackage{pifont}
\usepackage{enumitem}
\usepackage[hidelinks]{hyperref}

\begin{document}

\title{Avoiding Structural Failure Modes in Tabular Fair SSL: Online Primal-Dual Allocation under Confidence Gating}
\titlerunning{Avoiding Structural Failure Modes in Tabular Fair SSL}

\author{Hangchuan Liang\textsuperscript{\rm 1} and Changchun Li\textsuperscript{\rm 1,\rm 2}\thanks{Corresponding author.}}
\institute{\textsuperscript{\rm 1}College of Computer Science and Technology, Jilin University, China\\
    \textsuperscript{\rm 2}Key Laboratory of Symbolic Computation and Knowledge Engineering of the Ministry of Education,
Jilin University, China}

\maketitle

\begin{abstract}
Semi-supervised learning (SSL) enables prediction with limited labels, but high-stakes tabular applications (medical, credit, recidivism) require statistical fairness guarantees.
We identify a structural conflict in tabular fair SSL through a diagnostic stress test: under confidence-gated pseudo-labeling, moment-matching fairness regularizers can trigger two failure modes---Masking Collapse (fairness erodes confidence, starving pseudo-labels) and Trivial Saturation (drift to constant predictors).
We propose Online Primal-Dual Allocation (OPDA), an online controller that schedules fairness and entropy-based stability penalties using violation, risk, and pseudo-label health signals, avoiding per-dataset selection of a fixed fairness weight within this diagnostic regime.
On the evaluated tabular benchmarks (Adult, ACSIncome, COMPAS), OPDA mitigates the degenerate regimes observed under static weighting and simple single-signal adaptive baselines.
On Adult and COMPAS, it yields non-degenerate operating points competitive with the empirical static-$\lambda$ frontier; on ACSIncome, it preserves utility with a wider fairness-utility spread.
Relative to OPDA-lite, the full controller mainly shifts the operating point toward higher utility on ACSIncome, while Adult highlights the fairness--utility trade-off between the two variants.
These results position OPDA as a calibration-free controller for non-degenerate operating points in tabular fair SSL without per-dataset tuning.

\end{abstract}

\section{Introduction}
\label{sec:introduction}

Consider high-stakes tabular applications such as medical decision support, credit scoring, and recidivism prediction, where labeled data is scarce due to expensive expert annotations, yet regulatory compliance mandates statistical fairness guarantees across demographic groups.
Semi-supervised learning (SSL) enables prediction with limited labels, but modern pseudo-label SSL relies on \emph{confidence gating}: pseudo-labels are retained only when predictions exceed threshold $\tau$ \cite{fixmatch_ref}.
However, fairness regularizers enforce moment matching across groups, systematically suppressing confidence and disabling the pseudo-labeling mechanism.

\paragraph{Scope.}
This work focuses on \emph{tabular multi-output} fair SSL, motivated by three factors:
(1) \textbf{Prevalence in high-stakes applications}: tabular data dominates medical decision support, credit scoring, and recidivism prediction, where both label scarcity and fairness constraints are critical \cite{adult_ref,folktables_ref,propublica_compas_ref};
(2) \textbf{Isolation of failure mechanisms}: the multi-label structure ($L{>}1$) allows us to isolate and verify the predicted failure modes without confounding factors from high-dimensional representations;
(3) \textbf{Computational efficiency}: tabular benchmarks enable extensive ablations and sensitivity analyses that would be prohibitively expensive on image data.
Accordingly, our empirical evaluation uses a diagnostic stress-test setting to isolate and verify the failure modes, but the identified mechanisms (Masking Collapse, Trivial Saturation) are fundamental to confidence-gated fair SSL, and OPDA's multi-signal design is applicable to standard single-label settings.

\begin{figure}[t]
\centering
\includegraphics[width=0.85\columnwidth]{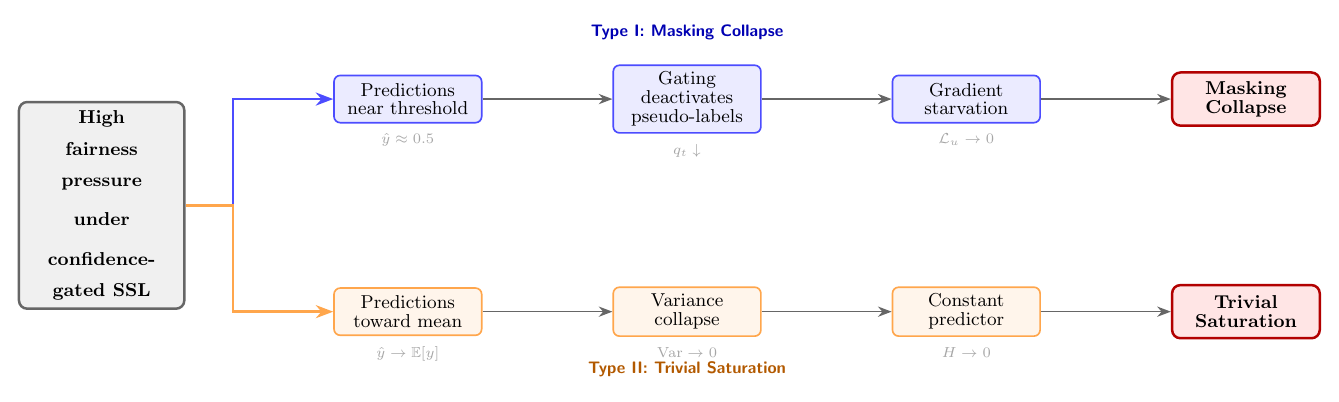}
\caption{\textbf{Causal mechanisms of structural failures in fair SSL.}
High fairness pressure diverges into two pathways:
\textbf{Type I (Masking Collapse)} erodes confidence and starves gradients;
\textbf{Type II (Trivial Saturation)} collapses predictions to constants with degraded utility.}
\label{fig:failure_mechanism}
\end{figure}

We identify two \textbf{structural failure modes}.
\textbf{Type~I (Masking Collapse):} fairness pressure concentrates predictions near decision boundaries, pseudo-label coverage $q_t$ collapses, yielding gradient starvation.
\textbf{Type~II (Trivial Saturation):} training drifts toward constant predictors with near-zero fairness violation but severely degraded utility.
Both failures arise from \emph{non-stationary} coupling between pseudo-label selection and constraint evaluation, rendering static weighting brittle.
Figure~\ref{fig:failure_mechanism} illustrates these causal pathways.

We formalize these pathologies from three perspectives.
First, under confidence-gated SSL, sufficiently strong moment-matching penalties admit constant solutions with vanishing gradients (Proposition~1; Appendix~A).
Second, for SimFair \cite{simfair_ref}, fairness enforcement is strictly sign-conflicting with entropy-based confidence sharpening in logit space (Proposition~2; Appendix~B).
Third, we provide convex-surrogate analysis of OPDA's outer-loop dynamics with sublinear-regret guarantees (Appendix~G).

This distinguishes our approach from multi-objective methods (PCGrad \cite{pcgrad_ref}, MGDA \cite{mgda_ref}) that assume stationary objectives.
Fair SSL under confidence gating exhibits endogenous non-stationarity: constraint functions $V_t(\theta)$ and $H_t(\theta)$ depend on evolving pseudo-label mask $M_t(\theta)$.
OPDA tracks a moving equilibrium by adapting dual weights epoch-by-epoch using online observables.

We propose \textbf{Online Primal-Dual Allocation (OPDA)} to resolve this structural deadlock.
OPDA uses bilevel budget--allocation parameterization: $\lambda_v^{(t)} = B_t \pi_t$ and $\lambda_h^{(t)} = B_t(1-\pi_t)$.
We update $(B_t,\pi_t)$ using online observables including fairness violation ($v_t$), risk proxy ($r_t$), pseudo-label health signals ($q_t,p_t,\mathrm{ESS}_t$), and gradient alignment.
Budget dynamics implement noise-robust knee-seeking; allocation dynamics are conflict-aware and anti-starvation-guaranteed.
OPDA runs with the same default configuration across all datasets (supplementary Appendix~D).

\paragraph{Contributions.}
\begin{itemize}[leftmargin=*,nosep]
    \item \textbf{Mechanism diagnosis.} We identify two failure modes in confidence-gated tabular fair SSL and formalize sufficient local mechanisms: vanishing-gradient neighborhoods (Proposition~1; Appendix~A) and exact logit-space sign conflict (Proposition~2; Appendix~B).

    \item \textbf{Pseudo-label-health-aware controller.} We propose OPDA, an epoch-level budget-allocation controller that schedules fairness and stability pressure using online observables, with a single default configuration across the evaluated tabular settings.

    \item \textbf{Empirical evidence in tabular stress tests.} Across Adult, ACSIncome, and COMPAS, OPDA mitigates degenerate regimes observed under static weighting and naive single-signal controllers, while producing competitive operating points in a diagnostic stress-test setting.
\end{itemize}

\paragraph{Code availability.}
{\sloppy
Code is available at\\
\url{https://anonymous.4open.science/r/OPDA-BB0C}.%
\par}

\section{Related Work}

\paragraph{Semi-Supervised Learning.}
Modern SSL uses consistency regularization with confidence-gated pseudo-labeling \cite{fixmatch_ref,flexmatch_ref,softmatch_ref}.
The unsupervised loss is masked when predictions fall below threshold $\tau$, introducing switch-like dependency.
In multi-label settings, gating is applied element-wise \cite{multilabel_ssl_ref1,multilabel_ssl_ref2}.
Most SSL methods assume external regularizers do not systematically disable the unlabeled objective.
Our analysis reveals moment-matching pressure can suppress confidence and deactivate pseudo-label masks.

\paragraph{Statistical Fairness via Moment Matching.}
Group fairness notions (DP, EOp) \cite{zafar_fairness_ref,hardt_equality_ref} control disparities across sensitive groups.
Differentiable surrogates use moment matching or distributional discrepancy \cite{mmd_fairness_ref,fair_moment_matching_ref}.
We adopt SimFair \cite{simfair_ref} as representative instantiation.
Crucially, constraint evaluation and pseudo-label selection share evolving predictions, making constraint functions \emph{endogenously non-stationary}.
OPDA schedules dual pressure using online observables rather than assuming fixed constraint landscape.

\paragraph{Fairness in Semi-Supervised Learning.}
Prior fair SSL work uses primal--dual updates \cite{fair_ssl_dual_ref}, group-aware re-weighting \cite{fair_ssl_reweight_ref,fair_pseudolabel_calibration_ref}, or representation-level fairness \cite{fair_repr_adv_ref,fair_invariant_repr_ref}.
These approaches focus on reducing bias given pseudo-labels.
We address an orthogonal concern: strong moment-matching pressure can destabilize pseudo-labeling itself, yielding Masking Collapse or Trivial Saturation.

\paragraph{Dynamic Weighting and Feedback Control.}
Multi-objective methods (GradNorm \cite{gradnorm_ref}, PCGrad \cite{pcgrad_ref}, MGDA \cite{mgda_ref}) adjust weights based on gradient statistics.
However, they operate on gradients alone and do not monitor mechanism health.
In fair SSL, pseudo-label availability can degrade catastrophically when fairness pressure suppresses confidence.
OPDA operates at epoch granularity and explicitly tracks pseudo-label health signals ($q_t, p_t, \mathrm{ESS}_t$) to detect SSL mechanism failure before gradient pathology.

We frame OPDA as feedback control \cite{control_theory_ml_ref,online_control_ml_ref} that tracks a moving equilibrium (Proposition~3; Appendix~C of the supplementary material).
OPDA separates total dual budget from allocation, updating both using online observables.
This provides a bridge to online-optimization interpretation with regret guarantees (Appendix~G).

\section{The OPDA Framework}
\label{sec:opda}

To resolve the structural deadlock between moment-matching fairness and confidence-gated pseudo-labeling, we propose \textbf{Online Primal-Dual Allocation (OPDA)}.
OPDA treats each epoch as an online round and adaptively schedules two dual weights: fairness penalty and entropy-based stability penalty, using online observables ($v_t, r_t, q_t, p_t, \mathrm{ESS}_t$, gradient alignment).

\paragraph{Why entropy-based stability?}
FixMatch-style confidence gating retains pseudo-labels only when predictions are sufficiently confident \cite{fixmatch_ref}.
To preserve pseudo-labeling feasibility, we introduce teacher-view entropy minimization as a classical SSL regularizer \cite{entropy_min_ref}.
OPDA does not assume this channel is always beneficial: its allocation weight can be driven to a small floor when mask starvation is not dominant.

\paragraph{Signals.}
OPDA monitors multiple online observables to detect both fairness violations and pseudo-label health degradation.
At epoch $t$: $v_t$ is training-time fairness penalty (SimFair DP-style moment matching, Appendix~E);
$r_t = 1-\mathrm{MacroF1}_{\mathrm{val}}\in[0,1]$ is risk proxy;
$q_t\in[0,1]$ is pseudo-label pass ratio;
$p_t\in[0,1]$ is proxy accuracy;
$\mathrm{ESS}_t>0$ is effective sample size.
We track pseudo-label health due to known self-training pathologies \cite{conf_bias_ref}.

\paragraph{Risk proxy as an outer-loop monitor.}
We define $r_t = 1-\mathrm{MacroF1}_{\mathrm{val}}$ on a held-out validation split as a bounded scalar that tracks utility degradation under class imbalance.
The validation Macro-F1 used in $r_t$ is binarized with a fixed threshold of $0.5$ during training; final reported test Macro-F1 uses the per-label rescaling thresholds described in Sec.~\ref{sec:experiments}.
Crucially, $r_t$ is \emph{not} added to the inner-loop training objective, and we do \emph{not} backpropagate through Macro-F1; it is used only by OPDA's outer-loop controller to gate budget growth against excessive utility loss, avoiding overfitting to validation performance.

\paragraph{Fairness notion.}
Optimized constraint $V_t(\theta)$ is DP-style group moment matching (Appendix~E) \cite{zafar_fairness_ref}.
We use \textbf{DP gap} as primary fairness metric; EOp/EOd are evaluation-only metrics \cite{hardt_equality_ref}.

\subsection{Fair SSL as Non-stationary Constrained Optimization}
\label{sec:opda_formulation}

Let $F_t(\theta)$ denote base SSL objective in epoch $t$:
\begin{equation}
F_t(\theta) = \mathcal{L}_{\mathrm{sup}}(\theta)+\lambda_u\,\mathcal{L}_{\mathrm{unsup}}\!\big(\theta; M_t(\theta)\big),
\end{equation}
where $M_t(\theta)$ is selection mask under confidence-thresholding \cite{fixmatch_ref}.
We consider two constraint functions: fairness violation $V_t(\theta)$ (SimFair DP) and entropy-based stability $H_t(\theta)$ on weak views.
Crucially, both are non-stationary through dependence on $M_t(\theta)$:
\begin{equation}
V_t(\theta)=V(\theta; M_t(\theta)),\qquad H_t(\theta)=H(\theta; M_t(\theta)).
\end{equation}

Proxy constrained formulation:
\begin{align}
\min_{\theta}\quad & F_t(\theta)\nonumber\\
\text{s.t.}\quad & V_t(\theta)\le \epsilon_v,\qquad H_t(\theta)\le \epsilon_h,
\end{align}
with time-varying Lagrangian
\begin{equation}
\mathcal{L}_t(\theta,\lambda_v^{(t)},\lambda_h^{(t)})
= F_t(\theta) + \lambda_v^{(t)}V_t(\theta)
+ \lambda_h^{(t)}H_t(\theta).
\end{equation}
OPDA avoids static fairness weight via validation; instead, it steers dual variables using online improvement-vs-risk signals.

\subsection{Dual Re-parameterization: Budget and Allocation}
\label{sec:opda_reparameterization}

OPDA re-parameterizes dual variables into \textbf{total budget} $B_t$ and \textbf{allocation ratio} $\pi_t$:
\begin{align}
\lambda_v^{(t)} &= B_t\,\pi_t,\qquad
\lambda_h^{(t)} = B_t\,(1-\pi_t), \\
\text{where } & B_t\in[B_{\min},B_{\max}^{\mathrm{soft}}],\quad \pi_t\in[0,1]. \nonumber
\end{align}
This decouples enforcement intensity ($B_t$) from conflict resolution ($\pi_t$).
In the released implementation, the log-domain update uses a numerical floor $\varepsilon_B$ when $B_{\min}=0$, and the non-binding cap is implemented as $B_{\max}^{\mathrm{soft}}$ (Table~\ref{tab:opda_defaults}).

\begin{figure}[t]
\centering
\includegraphics[width=0.95\columnwidth]{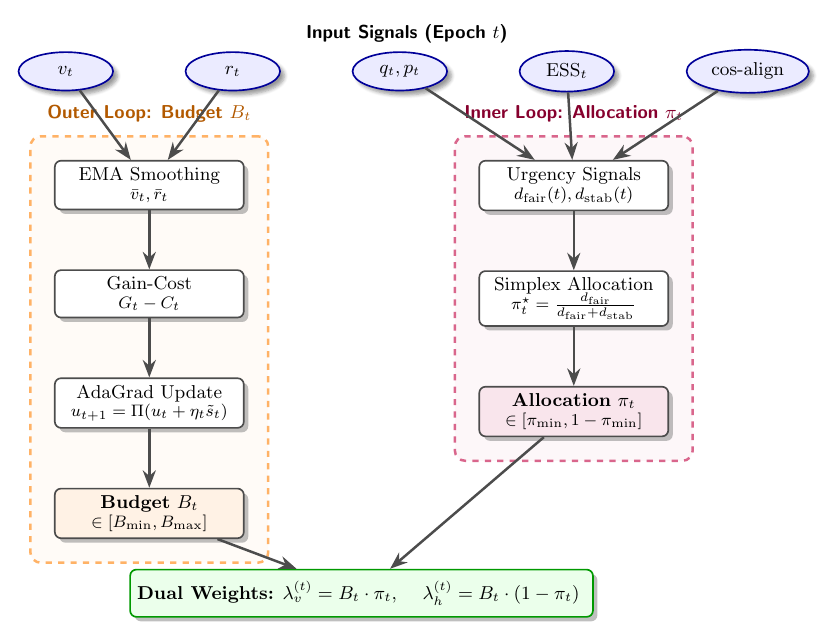}
\caption{\textbf{OPDA architecture.} Outer loop updates total budget $B_t$ using gain-cost signals; inner loop allocates budget between fairness and stability via alignment-gated urgency.}
\label{fig:opda_mechanism}
\end{figure}

OPDA uses fixed internal constants across datasets (Table~\ref{tab:opda_defaults}; supplementary Appendix~D).
Figure~\ref{fig:opda_mechanism} illustrates signal flow.

\subsubsection{Knee-Seeking Budget Dynamics}
\label{sec:opda_budget}

We update total enforcement budget once per epoch in the numerically stabilized log domain
\(
u_t=\log\!\big(\max(B_t,\;B_{\min}+\varepsilon_B)\big)
\).
Let $\bar v_t,\bar r_t$ denote EMA-smoothed fairness penalty and risk proxy.
OPDA forms signed improvement-vs-risk signal
\begin{align}
s_t &= G_t - C_t, \\
G_t &=\big[-\Delta \bar v_t - m_v(t)\big]_+, \\
C_t &=\big[\Delta \bar r_t - m_r(t)\big]_+,
\end{align}
where $m_v(t)$ and $m_r(t)$ are noise margins (supplementary Appendix~D).
Thus, $s_t>0$ only when fairness improves beyond noise without risk deterioration beyond noise.
The controller applies stabilization to obtain $\tilde{s}_t$ and updates
\begin{align}
u_{t+1} &= \Pi_{\mathcal U}\!\Big(u_t + \eta_t\,\tilde{s}_t\Big), \\
B_{t+1} &= \Pi_{[B_{\min},B_{\max}^{\mathrm{soft}}]}\!\big(\exp(u_{t+1})\big).
\end{align}

\subsubsection{Conflict-Aware Allocation}
\label{sec:opda_allocation}

OPDA allocates budget $B_t$ between penalties using allocation ratio $\pi_t\in[0,1]$.
Let $g_{\mathrm{base}}=\nabla_\theta F_t(\theta)$, $g_v=\nabla_\theta V_t(\theta)$, $g_h=\nabla_\theta H_t(\theta)$.
OPDA computes alignment cosines
\begin{equation}
c_t^{v}=\cos(g_v,g_{\mathrm{base}}),\qquad c_t^{h}=\cos(g_h,g_{\mathrm{base}}),
\end{equation}
and uses sigmoid gate $\phi(c)$ to suppress urgency when constraint gradient is anti-aligned.

OPDA forms two urgency scalars $d_v(t)$ and $d_h(t)$.
$d_v(t)$ summarizes noise-filtered fairness improvement and current violation;
$d_h(t)$ summarizes pseudo-label health deficits ($q_t, p_t, \mathrm{ESS}_t$) relative to warmup baselines.
Both are EMA-smoothed and gated by $\phi(\cdot)$ (supplementary Appendix~D).

Target allocation minimizes convex quadratic surrogate:
\begin{align}
\pi_t^{\star} &= \arg\min_{\pi\in[0,1]} \Big( d_v(t)(\pi-1)^2+d_h(t)\pi^2 \Big) \\
&= \frac{d_v(t)}{d_v(t)+d_h(t)+\varepsilon}.
\end{align}
To prevent starvation, OPDA enforces ESS-adaptive floor $\pi_{\min}(t)>0$ with EMA and clipping:
\begin{align}
\pi_t &= \Pi_{[\pi_{\min}(t),\,1-\pi_{\min}(t)]}\Big(\mathrm{EMA}(\pi_t^{\star})\Big), \nonumber \\
\lambda_v^{(t)} &= B_t\pi_t,\quad \lambda_h^{(t)}=B_t(1-\pi_t).
\end{align}

\subsection{Formal Characterization of Failure Modes}
\label{sec:failure_theory}

We formalize the two structural failures identified in Sec.~\ref{sec:introduction}.

\paragraph{Scope.}
The following propositions analyze failure modes under convex surrogate losses.
In practice, OPDA operates on non-convex deep networks.
We view these propositions as providing qualitative guidance---they identify structural pathologies (vanishing gradients, gradient conflict) that motivate OPDA's multi-signal design.
Empirical validation is provided in Sec.~\ref{sec:experiments}.

\begin{proposition}[Masking Collapse: vanishing-gradient neighborhoods]
\label{prop:masking_collapse}
Consider confidence-gated SSL with threshold $\tau\in(1/2,1)$ and moment-matching fairness penalty $V(\theta)$ (e.g., squared DP).
Suppose the model can realize constant predictors, and let $\theta^*$ satisfy $f_{\theta^*}(x)\equiv c$ with $V(\theta^*)=0$.
If $c\in(1-\tau,\tau)$ (low-confidence region), then there exists open neighborhood $\mathcal{N}(\theta^*)$ such that for all $\theta\in\mathcal{N}(\theta^*)$, $\mathcal{L}_{\mathrm{unsup}}(\theta)=0$ and $\nabla_\theta \mathcal{L}_{\mathrm{unsup}}(\theta)=0$.
\end{proposition}

\noindent\textbf{Intuition.}
Strong fairness pressure drives predictions toward decision boundary ($\hat{y} \approx 0.5$), causing confidence gate to reject most samples.
This creates vanishing-gradient neighborhoods preventing learning from unlabeled data.
Proof in Appendix~A.

\begin{proposition}[Gradient conflict under SimFair]
\label{prop:gradient_conflict}
Consider binary classification with two groups ($K=2$).
Let $V_{\mathrm{SimFair}}(\theta) = \|\mu_0(\theta)-\mu_1(\theta)\|_2$ denote SimFair DP penalty, and $H(\theta)$ denote entropy minimization.
Assume: (i) imbalanced groups, (ii) distinct group means, (iii) differentiable model.
Then logit-space gradients exhibit sign conflict: fairness pushes toward mean equalization while entropy minimization pushes predictions away from $0.5$.
Whether this translates to parameter-space negative alignment depends on network Jacobian and is monitored online via cosine alignment in OPDA.
\end{proposition}

\noindent\textbf{Intuition.}
SimFair's moment-matching reduces $\|\mu_0-\mu_1\|$ (reducing inter-group variance).
Entropy minimization drives predictions toward $\{0,1\}$ (increasing prediction variance).
Under imbalanced groups, these create persistent gradient conflict.
Proof in Appendix~B.

\subsection{Theoretical Justification (Sketch)}
\label{sec:opda_theory}

We emphasize these statements are theory on convex surrogates; they do not claim global optimality for deep non-convex training.

\begin{remark}[Outer-loop regret interpretation]
OPDA's budget update in log domain $u_t=\log B_t$ can be viewed as projected online gradient descent on convex surrogate loss.
Standard OGD analysis \cite{zinkevich_oco_ref,hazan_oco_ref} with step size $\eta=\Theta(1/\sqrt{T})$ yields $O(\sqrt{T})$ regret.
This provides theoretical foundation for adaptive scheduling, though non-convex inner loop means we do not claim end-to-end optimality.
\end{remark}

\begin{lemma}[Anti-starvation of dual weights]
\label{lem:antistarvation}
Let $\underline{B}>0$ denote effective budget lower bound, and assume $B_t\ge \underline{B}$ and $\pi_t\in[\pi_{\min}(t),1-\pi_{\min}(t)]$ with $\pi_{\min}(t)>0$.
Then at every epoch, $\lambda_v^{(t)} \ge \underline{B}\,\pi_{\min}(t)$ and $\lambda_h^{(t)} \ge \underline{B}\,\pi_{\min}(t)$, so neither penalty can be fully turned off.
\end{lemma}

\noindent\textbf{Empirical validation.}
Figure~\ref{fig:dynamic_trajectories} (Sec.~\ref{sec:experiments}) shows maintaining $\lambda_h^{(t)} > 0$ correlates with pseudo-label coverage $q_t$ remaining above critical threshold.

\begin{remark}[Knee-seeking budget dynamics]
\label{rem:steady_state}
Suppose EMA-smoothed signals stabilize.
Then OPDA's steady state satisfies: (i) improvement-vs-risk signal vanishes ($\tilde{s}_t \approx 0$); (ii) further budget increases improving fairness beyond noise are associated with risk increases beyond noise.
Empirically, this is consistent with Adult and COMPAS ending near the favorable region of the static-$\lambda$ frontier (Figure~\ref{fig:frontier}), while ACSIncome remains in a broader interior region.
\end{remark}

\noindent\textbf{Scope.}
This characterizes OPDA's controller-level equilibrium under deterministic drift approximation (Appendix~G).
It does not claim global optimality but provides precise interpretation of knee-seeking behavior in controller units.

\section{Experiments}
\label{sec:experiments}

We evaluate OPDA under semi-supervised training with SimFair regularization \cite{simfair_ref} to answer:
\textbf{(RQ1) Diagnosis:} Do the two structural failure modes manifest under fixed dual weighting?
\textbf{(RQ2) Trade-off:} Can OPDA produce competitive utility--fairness operating point without manual tuning?
\textbf{(RQ3) Mechanism:} Do induced dual schedules evolve consistently with OPDA's design?
\textbf{(RQ4) Comparison:} Does OPDA's multi-signal design offer advantages over single-signal adaptive baselines?

\subsection{Experimental Settings}

\paragraph{Diagnostic design rationale.}
Our experimental setting is a \textbf{diagnostic stress test} designed to isolate and expose pseudo-label-health failures under shared fairness pressure.
On Adult, we construct a 23-dimensional label vector (income + workclass + occupation one-hot blocks) and apply fairness constraints to all dimensions.
This design choice serves two purposes:
(i) \textbf{Correlated attributes}: \texttt{workclass} and \texttt{occupation} are predictive of \texttt{income} and may themselves exhibit demographic disparities, so constraining only \texttt{income} could allow bias to propagate through auxiliary predictions.
(ii) \textbf{Multi-label stress test}: Applying fairness constraints to all dimensions creates a more challenging optimization landscape that exercises OPDA's conflict-aware allocation mechanism.
Standard single-label deployment would use simpler configurations; we adopt this multi-label structure specifically to stress-test the controller's ability to avoid structural failures.
To verify that improvements are not artifacts of this multi-label aggregation, we report per-dimension decomposition metrics (Tables~\ref{tab:decomp_adult_compact} and~\ref{tab:decomp_acs_compact}) that isolate the primary income prediction task.

\paragraph{Datasets.}
\textbf{Adult} \cite{adult_ref}: sensitive attribute \texttt{sex} ($K{=}2$), $L{=}23$ dimensions (income + workclass + occupation one-hot blocks).
\textbf{ACSIncome} \cite{folktables_ref}: sensitive attribute \texttt{RAC1P} ($K{=}9$ race groups), $L{=}4$ binary targets (income\_over\_50k, employed, pubcov, migrated).
\textbf{COMPAS} \cite{propublica_compas_ref}: sensitive attribute \texttt{sex} ($K{=}2$), $L{=}2$ binary targets (general\_recid, violent\_recid).

\paragraph{Metrics.}
\textbf{Macro-F1} (macro-averaged over $L$ dimensions) as primary utility metric, using per-label rescaling thresholds (empirical prevalence).
\textbf{DP gap} as primary fairness metric: $\mathrm{DP\ gap} \triangleq \sum_{k=1}^{K}\left\lVert \mathbb{E}[\hat{y}] - \mathbb{E}[\hat{y}\mid a{=}k]\right\rVert_2$.
We additionally report \textbf{EOp gap} for completeness.
Per-dimension decomposition metrics in Tables~\ref{tab:decomp_adult_compact} and~\ref{tab:decomp_acs_compact} isolate primary income prediction.
Those decomposition tables additionally report \textbf{Binary EOd} on the primary task, so the main tables and the decomposition tables use different secondary fairness diagnostics.

\paragraph{Methods.}
\textbf{FixMatch (Base)}: Standard SSL \cite{fixmatch_ref}.
\textbf{Static-$\lambda$}: FixMatch + SimFair with fixed $\lambda_{\mathrm{fair}}$.
We run grid $\lambda_{\mathrm{fair}} \in \{0.1, 1, 10, 20, \ldots, 100\}$ as empirical frontier.
\textbf{Adaptive baselines}: Three single-signal controllers (EMA-P, PI, DualAsc) that update $\lambda_{\mathrm{fair}}$ once per epoch using only the EMA-smoothed fairness violation $\bar{v}_t$ (with momentum $\rho{=}0.9$, matching OPDA's EMA).
Unlike OPDA, none monitors pseudo-label health ($q_t, p_t, \mathrm{ESS}_t$), gradient alignment, or risk proxy—this isolation tests whether fairness-only feedback suffices to avoid structural failures.
All three use a single pre-specified configuration with the target violation $v_{\mathrm{tgt}}$ calibrated deterministically from warmup-epoch violations (no tuning required; supplementary Appendix~I).
Single-signal controllers face a fundamental dilemma: they cannot distinguish between fairness violations caused by insufficient model capacity (requiring higher $\lambda_{\mathrm{fair}}$) versus pseudo-label starvation (requiring lower $\lambda_{\mathrm{fair}}$), leading to oscillation or premature saturation.
\textbf{OPDA}: Schedules $(\lambda_v^{(t)},\lambda_h^{(t)})$ via $(B_t,\pi_t)$ with fixed constants (Table~\ref{tab:opda_defaults}).
\textbf{MOO baselines}: PCGrad \cite{pcgrad_ref} and CAGrad \cite{cagrad_ref} for gradient-based comparison.

\begin{table}[t]
\centering
\small
\setlength{\tabcolsep}{5pt}
\renewcommand{\arraystretch}{1.08}
\caption{\textbf{Fixed OPDA controller configuration used throughout the paper (pre-specified; not tuned).}
These quantities are \emph{internal constants} of the controller implementation and are held fixed across datasets/backbones unless explicitly perturbed for diagnostics in supplementary Appendix~H. The released implementation additionally uses a numerical log-floor $\varepsilon_B=10^{-8}$ in the log/exp budget parameterization to keep $u_t$ well-defined even when $B_{\min}=0$.}
\label{tab:opda_defaults}
\label{tab:opda_defaults_full}
\begin{tabular}{l l}
\toprule
\textbf{Internal constant} & \textbf{Default} \\
\midrule
EMA momentum $\rho$ & $0.9$ \\
Warmup length $T_{\mathrm{warm}}$ (epochs) & $20$ \\
Rolling window $W=\mathrm{clip}(T_{\mathrm{warm}},10,30)$ & $20$ \\
Noise margin $k$ in gain/cost filters & $1.0$ \\
Allocation-gate sharpness $\beta$ in $\phi(c)=\sigma(\beta c)$ & $8.0$ \\
Budget clip floor $B_{\min}$ (code default) & $0.0$ \\
Numerical log-floor $\varepsilon_B$ (stabilization) & $10^{-8}$ \\
Soft budget cap $B_{\max}^{\mathrm{soft}}$ (non-binding default) & $10^{6}$ \\
Anti-starvation floor $\pi_{\min}(t)$ & $\frac{1}{10+2\max(1,\mathrm{ESS}_t)}$ \\
Leak coefficient $\beta_{\mathrm{leak}}(t)$ & $\frac{1}{\sqrt{1+t}}$ \\
\bottomrule
\end{tabular}
\end{table}

\subsection{Main Results}

\begin{figure}[t]
\centering
\includegraphics[width=\columnwidth]{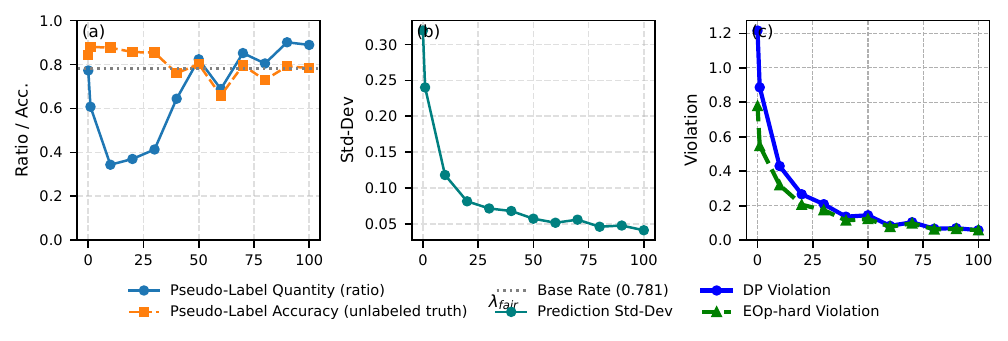}
\caption{\textbf{Illustrative failure modes under excessive fairness pressure.}
\textbf{Adult:} Type I---coverage collapse under static $\lambda_{\mathrm{fair}}$.
\textbf{ACSIncome:} Type II---constant predictor drift under aggressive single-signal control.}
\label{fig:anatomy}
\end{figure}

\paragraph{RQ1: Failure mode diagnosis.}
Figure~\ref{fig:anatomy} illustrates two structural failures under excessive fairness pressure.
On Adult, Static-$\lambda{=}100$ exhibits \textbf{Masking Collapse}: $q_t \to 0.02$, Macro-F1 $\to 0.15$ (near base-rate).
On ACSIncome, aggressive single-signal control can induce \textbf{Trivial Saturation}; supplementary Appendix~I shows PI reaches 100\% saturation for 5 of 6 target-violation settings across all five pseudo-labeling backbones.
Both failures validate Propositions~\ref{prop:masking_collapse} and~\ref{prop:gradient_conflict}.

\begin{figure*}[t]
\centering
\includegraphics[width=0.98\textwidth]{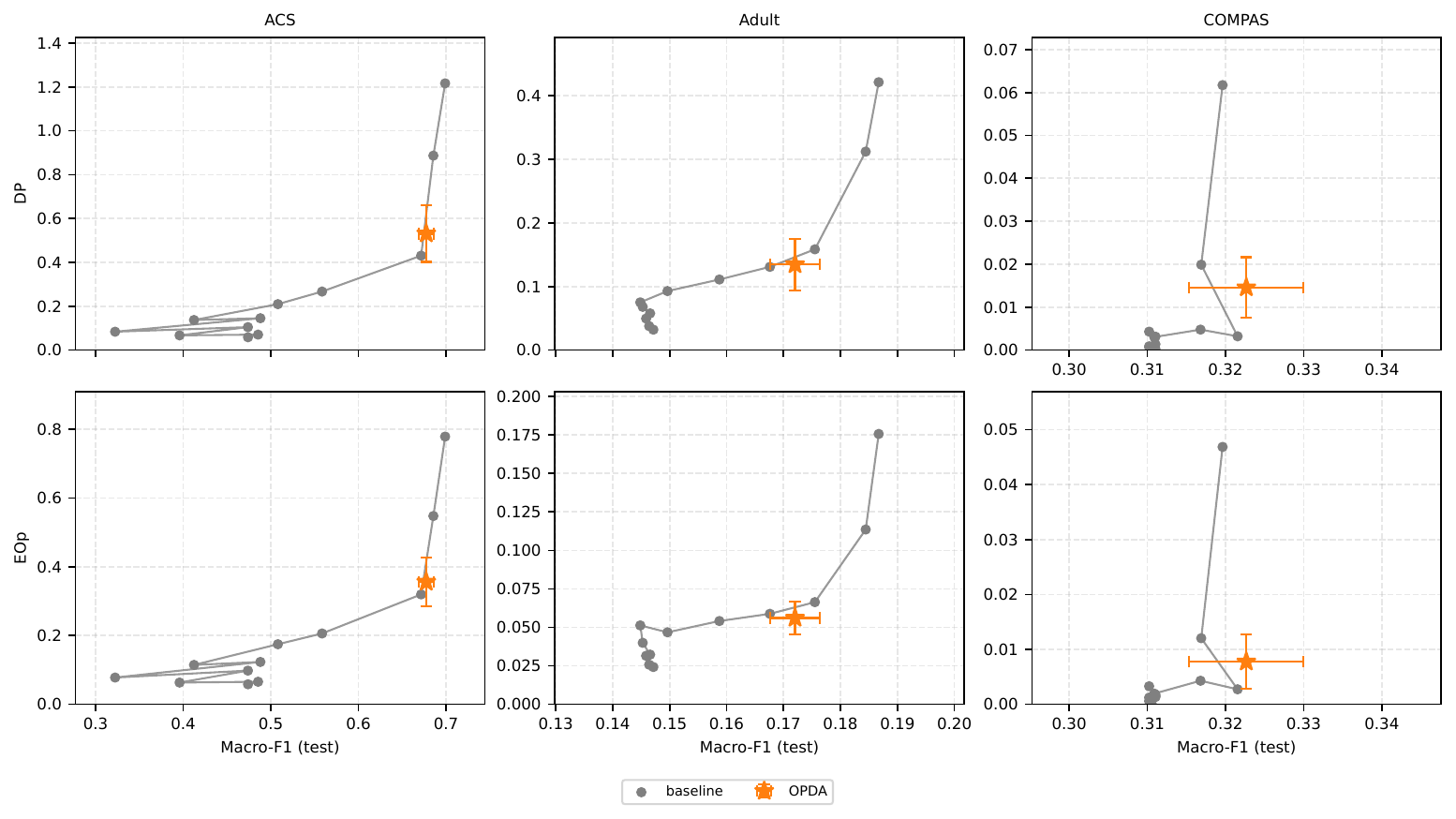}
\caption{\textbf{Empirical Pareto frontiers.} OPDA operating points (red stars) lie near the favorable region of the static-$\lambda$ frontier on Adult and COMPAS; broader spread on ACSIncome.}
\label{fig:frontier}
\end{figure*}

\paragraph{RQ2: Trade-off performance.}
Table~\ref{tab:main_results} summarizes results across three datasets.
On Adult and COMPAS, OPDA generates non-degenerate operating points that are competitive with the empirical static-$\lambda$ frontier (Figure~\ref{fig:frontier}).
On ACSIncome, OPDA preserves utility while exhibiting wider fairness-utility spread in the more complex $K{=}9$, $L{=}4$ setting, without guaranteeing uniform fairness improvement.
Decomposition metrics (Tables~\ref{tab:decomp_adult_compact}, \ref{tab:decomp_acs_compact}) show that the same utility--fairness trade-off appears on the primary income prediction task.

\begin{table}[t]
\centering
\small
\setlength{\tabcolsep}{4pt}
\caption{\textbf{Main results across three tabular benchmarks.}
Mean $\pm$ std over 10 randomly selected seeds. OPDA is positioned as a calibration-free operating-point generator rather than a guarantee of uniform dominance over every fixed static weight.}
\label{tab:main_results}
\resizebox{\columnwidth}{!}{
\begin{tabular}{llcccc}
\toprule
Dataset & Setting & Macro-F1 $\uparrow$ & Micro-F1 $\uparrow$ & DP $\downarrow$ & EOP $\downarrow$ \\
\midrule
Adult & Base & 0.184 $\pm$ 0.003 & 0.333 $\pm$ 0.008 & 0.383 $\pm$ 0.035 & 0.181 $\pm$ 0.022 \\
 & Static-$\lambda$ & 0.171 $\pm$ 0.002 & 0.300 $\pm$ 0.009 & 0.123 $\pm$ 0.016 & 0.063 $\pm$ 0.013 \\
 & OPDA & 0.165 $\pm$ 0.010 & 0.295 $\pm$ 0.051 & 0.113 $\pm$ 0.076 & 0.059 $\pm$ 0.028 \\
\midrule
ACSIncome & Base & 0.702 $\pm$ 0.008 & 0.797 $\pm$ 0.008 & 1.264 $\pm$ 0.116 & 0.797 $\pm$ 0.098 \\
 & Static-$\lambda$ & 0.663 $\pm$ 0.013 & 0.775 $\pm$ 0.012 & 0.365 $\pm$ 0.074 & 0.265 $\pm$ 0.043 \\
 & OPDA & 0.678 $\pm$ 0.006 & 0.797 $\pm$ 0.011 & 0.509 $\pm$ 0.043 & 0.338 $\pm$ 0.028 \\
\midrule
COMPAS & Base & 0.330 $\pm$ 0.009 & 0.586 $\pm$ 0.006 & 0.069 $\pm$ 0.005 & 0.056 $\pm$ 0.007 \\
 & Static-$\lambda$ & 0.319 $\pm$ 0.003 & 0.575 $\pm$ 0.002 & 0.004 $\pm$ 0.002 & 0.003 $\pm$ 0.001 \\
 & OPDA & 0.323 $\pm$ 0.005 & 0.576 $\pm$ 0.002 & 0.007 $\pm$ 0.002 & 0.003 $\pm$ 0.002 \\
\bottomrule
\end{tabular}
}
\end{table}

\begin{figure}[t]
\centering
\includegraphics[width=0.95\columnwidth]{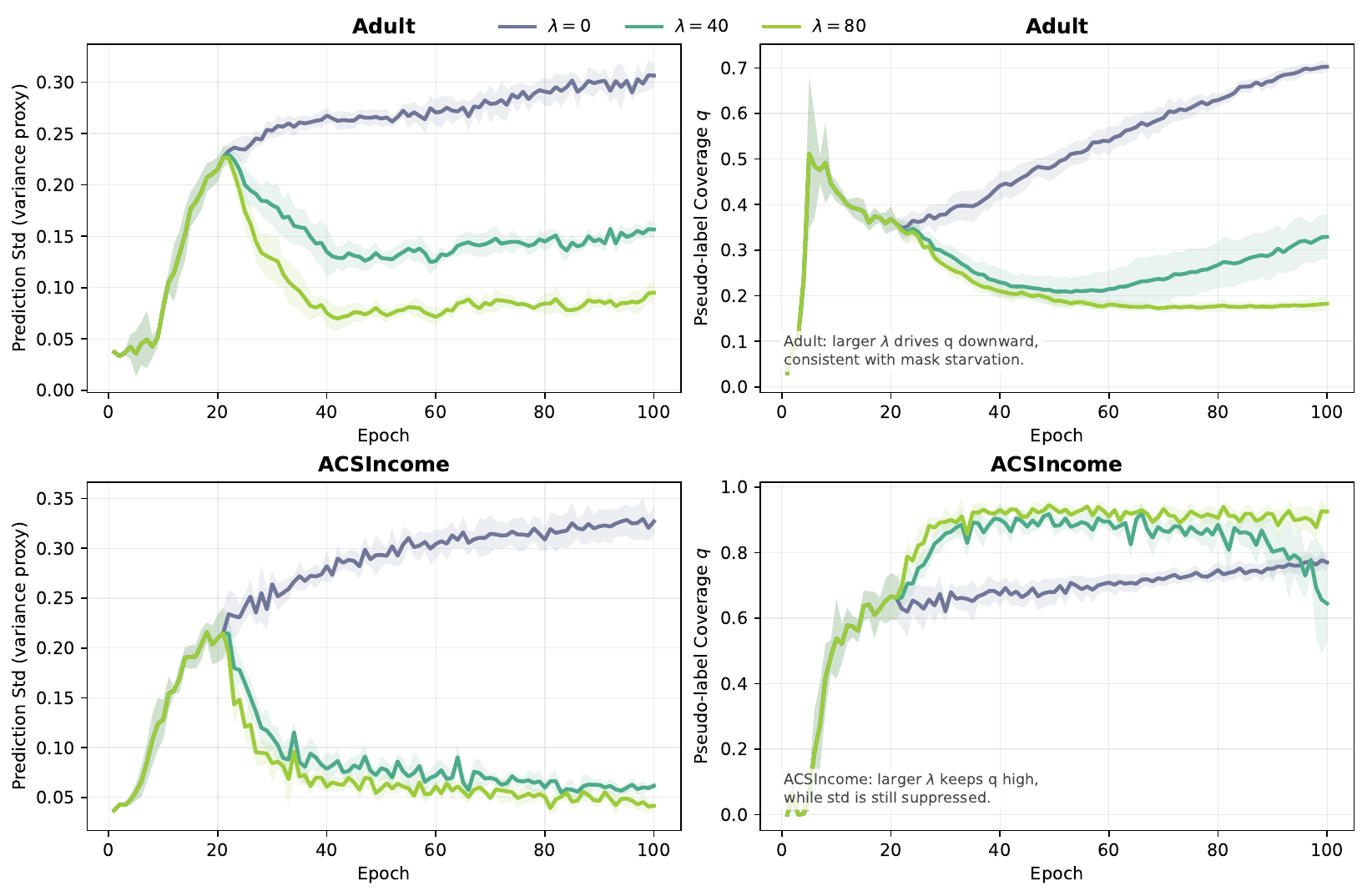}
\caption{\textbf{Failure-mechanism trajectories under representative fairness levels.}
Larger fairness pressure reduces pseudo-label coverage on Adult, while ACSIncome can retain high coverage even as prediction variance is suppressed.}
\label{fig:dynamic_trajectories}
\end{figure}

\paragraph{RQ3: Mechanism validation.}
Figure~\ref{fig:dynamic_trajectories} visualizes how representative fairness levels reshape prediction variance and pseudo-label coverage over training.
On Adult, larger fixed fairness pressure drives pseudo-label coverage $q_t$ downward after warmup, consistent with mask starvation.
On ACSIncome, coverage can remain high even while prediction variance is suppressed, consistent with constant-predictor drift.
These trajectories match the two failure mechanisms that motivate OPDA's non-zero stability channel.

\paragraph{RQ4: Comparison with adaptive baselines.}
EMA-P and DualAsc are reasonable single-signal baselines at the default target setting, but supplementary Appendix~I shows that their target-violation sensitivity is mild on Adult and much stronger on ACSIncome.
PI is the least robust baseline: supplementary Appendix~I shows 100\% saturation for 5 of 6 target settings across all five pseudo-labeling backbones.
DualAsc lacks OPDA's anti-starvation guarantee and pseudo-label-health monitoring.
OPDA's multi-signal design reduces saturation risk while maintaining competitive trade-offs.
Table~\ref{tab:moo_compact_main} adds a compact empirical comparison with gradient-based MOO.
On Adult, OPDA reaches lower DP than PCGrad/CAGrad, with utility comparable to CAGrad and below PCGrad; on ACSIncome, PCGrad/CAGrad recover utility but remain less fair; on COMPAS, gradient-only methods exhibit non-zero saturation, while OPDA remains at Sat.\% $= 0$ and is more fair than both PCGrad and CAGrad.
Table~\ref{tab:robustness_compact_main} summarizes ranges over five pseudo-labelers under one fixed configuration.
The controller remains in the same qualitative regime across backbones, so the observed behavior is not tied to a single SSL pipeline.

\begin{table}[t]
\centering
\scriptsize
\setlength{\tabcolsep}{3pt}
\caption{\textbf{Compact empirical comparison with gradient-based MOO methods} on the FixMatch backbone. Cells report mean Macro-F1 / DP / Sat.\%. Sat.\% is omitted for Static-$\lambda$ because it is a fixed-weight baseline. Full mean$\pm$std tables are provided in the supplementary PDF.}
\label{tab:moo_compact_main}
\resizebox{\columnwidth}{!}{
\begin{tabular}{lcccc}
\toprule
Dataset & Static-$\lambda$ & OPDA & PCGrad & CAGrad \\
\midrule
Adult & 0.171 / 0.123 / -- & \textbf{0.165 / 0.113 / 0} & 0.178 / 0.222 / 0 & 0.171 / 0.133 / 0 \\
ACSIncome & \textbf{0.663 / 0.365 / --} & 0.678 / 0.509 / 0 & 0.694 / 0.786 / 0 & 0.692 / 0.727 / 0 \\
COMPAS & \textbf{0.319 / 0.004 / --} & 0.323 / 0.007 / 0 & 0.317 / 0.016 / 20 & 0.316 / 0.011 / 30 \\
\bottomrule
\end{tabular}
}
\end{table}

\begin{table}[t]
\centering
\scriptsize
\setlength{\tabcolsep}{4pt}
\caption{\textbf{Cross-backbone ranges over five pseudo-labelers} (FixMatch, MeanTeacher, NoisyStudent, PseudoLabel, UDA). Cells report the range of mean Macro-F1 / mean DP obtained with a single fixed configuration per controller. This summarizes whether OPDA's operating regime depends on a specific SSL backbone.}
\label{tab:robustness_compact_main}
\resizebox{\columnwidth}{!}{
\begin{tabular}{lccc}
\toprule
Dataset & Base & Static-$\lambda$ & OPDA \\
\midrule
Adult & 0.184--0.186 / 0.380--0.424 & 0.167--0.171 / 0.120--0.139 & 0.165--0.170 / 0.113--0.160 \\
ACSIncome & 0.694--0.702 / 1.239--1.283 & 0.657--0.665 / 0.329--0.365 & 0.675--0.685 / 0.493--0.557 \\
COMPAS & 0.329--0.343 / 0.060--0.069 & 0.319--0.323 / 0.003--0.004 & 0.321--0.323 / 0.006--0.009 \\
\bottomrule
\end{tabular}
}
\end{table}

\subsection{Ablation and Sensitivity}

\paragraph{OPDA-lite ablation.}
OPDA-lite removes gradient alignment gating and replaces OPDA's full multi-signal budget logic with a simplified budget-and-allocation controller.
Relative to OPDA-lite, full OPDA shifts the operating point toward utility preservation when pseudo-label feasibility is fragile.
On Adult, the two methods achieve essentially the same Macro-F1, while OPDA-lite attains a lower DP gap.
On ACSIncome, full OPDA recovers higher Macro-F1 at the cost of a larger DP gap.
Detailed ablation results are provided in supplementary Appendix~I.

\paragraph{Sensitivity analysis.}
We evaluate robustness to controller constants: smoothing parameters ($\rho_{\mathrm{EMA}}$), noise margins ($m_v, m_r$), budget bounds ($B_{\min}, B_{\max}$), allocation floor ($\pi_{\min}$).
OPDA remains stable under moderate perturbations around defaults.
Detailed sensitivity results are provided in supplementary Appendix~H.

\subsection{Per-Dimension Decomposition}

Tables~\ref{tab:decomp_adult_compact} and~\ref{tab:decomp_acs_compact} report per-dimension metrics isolating primary income prediction.
In addition to per-dimension DP gap and F1, these tables report Binary DP and Binary EOd on the primary task.
On Adult income dimension: OPDA achieves F1 $= 0.69$, DP gap $= 0.15$ vs Static-$\lambda{=}10$ F1 $= 0.49$, DP gap $= 0.08$.
On ACS income dimension: OPDA achieves F1 $= 0.68$, DP gap $= 0.39$ vs Static-$\lambda{=}10$ F1 $= 0.66$, DP gap $= 0.23$.
These show that OPDA's utility-preserving operating point on the aggregate metrics is not an artifact of multi-label aggregation, while the utility--fairness trade-off remains visible on the primary task.

\begin{table}[t]
\centering
\scriptsize
\setlength{\tabcolsep}{3pt}
\caption{\textbf{Label-block decomposition on Adult (income dimension).}
Per-block DP gap and Macro-F1, plus Binary DP and Binary EOd on the primary income task.}
\label{tab:decomp_adult_compact}
\resizebox{\columnwidth}{!}{
\begin{tabular}{lcccc}
\toprule
Method & Income DP gap $\downarrow$ & Income F1 $\uparrow$ & Binary DP $\downarrow$ & Binary EOd $\downarrow$ \\
\midrule
Base & 0.279$\pm$0.012 & 0.675$\pm$0.022 & 0.337$\pm$0.025 & 0.265$\pm$0.020 \\
Static-10 & 0.082$\pm$0.006 & 0.485$\pm$0.045 & 0.070$\pm$0.018 & 0.019$\pm$0.010 \\
OPDA-full & 0.146$\pm$0.061 & \textbf{0.693$\pm$0.031} & 0.208$\pm$0.061 & 0.128$\pm$0.055 \\
\bottomrule
\end{tabular}
}

\end{table}

\begin{table}[t]
\centering
\scriptsize
\setlength{\tabcolsep}{2pt}
\renewcommand{\arraystretch}{1.05}
\caption{\textbf{Label-block decomposition on ACSIncome (target1 dimension).}
Target1 is the primary income prediction task; Binary DP and Binary EOd are reported on that task.}
\label{tab:decomp_acs_compact}
\resizebox{\columnwidth}{!}{
\begin{tabular}{lcccc}
\toprule
Method & Target1 DP gap $\downarrow$ & Target1 F1 $\uparrow$ & Binary DP $\downarrow$ & Binary EOd $\downarrow$ \\
\midrule
Base & 1.006$\pm$0.054 & 0.715$\pm$0.004 & 0.339$\pm$0.012 & 0.461$\pm$0.052 \\
Static-10 & 0.227$\pm$0.044 & 0.661$\pm$0.012 & 0.299$\pm$0.037 & 0.353$\pm$0.056 \\
OPDA-full & 0.387$\pm$0.080 & 0.677$\pm$0.009 & 0.280$\pm$0.040 & 0.406$\pm$0.054 \\
\bottomrule
\end{tabular}
}

\end{table}

\section{Conclusion}
\label{sec:conclusion}

We studied fair semi-supervised training under confidence gating and moment-matching regularization.
We identified two structural failure modes---Masking Collapse and Trivial Saturation---arising from non-stationary coupling between pseudo-label mask and fairness constraint.
We proposed OPDA, an online controller that adaptively schedules dual weights via budget--allocation reparameterization.
In tabular multi-output stress tests, OPDA provides calibration-free path to controlled operating points under label scarcity.
While our evaluation focuses on diagnostic stress tests to isolate failure mechanisms, the identified pathologies (Masking Collapse, Trivial Saturation) are inherent to confidence-gated fair SSL, and OPDA's design principles extend to standard single-label deployment scenarios.
While not universally superior across all metrics and datasets, OPDA's value lies in diagnosing and managing failure regimes without per-dataset tuning.
Comparisons with single-signal adaptive baselines provide additional evidence: PI exhibits trivial saturation across most target-violation settings, while EMA-P and DualAsc are comparatively robust on Adult but substantially more sensitive on ACSIncome.
Relative to OPDA-lite ablation, full OPDA mainly shifts the operating point toward higher utility on ACSIncome, whereas Adult highlights the fairness--utility trade-off between the two variants.

\paragraph{Limitations.}
Our study focuses on tabular multi-output benchmarks (Adult, ACSIncome, COMPAS).
Whether identified failure modes manifest in high-dimensional image/text fair-SSL remains open.
Adaptive baselines are intentionally minimal (single-signal); comparing against richer Pareto-based strategies is natural next step.
On ACSIncome, OPDA exhibits larger DP gap variance on multi-label aggregate, illustrating positioning as controller avoiding per-dataset weight selection rather than uniform fairness improvement guarantee.
OPDA requires pre-specifying controller constants such as warmup length $T_{\mathrm{warm}}$ and budget bounds, though we demonstrate robustness across a diagnostic range.

\paragraph{Future directions.}
(i) Extending OPDA to high-dimensional image/text benchmarks;
(ii) systematic comparison with Pareto-based adaptive strategies;
(iii) investigating dimension-weighted fairness variants where practitioners specify label-specific fairness relevance.

\paragraph{Use of Generative AI.}
We used generative AI tools only for language polishing and grammar correction. All technical content, experiments, analysis, and final wording were verified by the authors, who take full responsibility for the manuscript.

\bibliographystyle{splncs04}
\bibliography{reference}

\end{document}